\documentclass[conference]{IEEEtran}

\usepackage{graphicx,subcaption}
\usepackage{mathtools}
\usepackage{cases}
\usepackage{amsfonts}
\usepackage{verbatim}

\ifCLASSINFOpdf
\else
\fi
\begin{document}
%
\title{Neural Radiance Fields for Transparent Object Using Visual Hull}


\author{\IEEEauthorblockN{Heechan Yoon}
\IEEEauthorblockA{Dept. of Computer Science and Engineering\\
Kyunghee University\\
Yongin, Republic of Korea\\
harryyoon777@gmail.com
}
\and
\IEEEauthorblockN{Seungkyu Lee}
\IEEEauthorblockA{Dept. of Computer Science and Engineering\\
Kyunghee University\\
Yongin, Republic of Korea\\
seungkyu@khu.ac.kr}
}


%


\maketitle

\begin{abstract}
Unlike opaque object, novel view synthesis of transparent object is a challenging task, because transparent object refracts light of background causing visual distortions on the transparent object surface along the viewpoint change.
Recently introduced Neural Radiance Fields (NeRF) is a view synthesis method. Thanks to its remarkable performance improvement, lots of following applications based on NeRF in various topics have been developed.  
However, if an object with a different refractive index is included in a scene such as transparent object, NeRF shows limited performance because refracted light ray at the surface of the transparent object is not appropriately considered. 
To resolve the problem, we propose a NeRF-based method consisting of the following three steps: First, we reconstruct a three-dimensional shape of a transparent object using visual hull. Second, we simulate the refraction of the rays inside of the transparent object according to Snell's law. Last, we sample points through refracted rays and put them into NeRF. Experimental evaluation results demonstrate that our method addresses the limitation of conventional NeRF with transparent objects.
\end{abstract}

\begin{IEEEkeywords}
transparent objects, refraction, view synthesis
\end{IEEEkeywords}

%

\IEEEpeerreviewmaketitle

\section{Introduction and Related Works}
In our daily experiences, we often see that background behind transparent object appears distorted as it passes through the object due to refraction.
Light refraction occurring on the surface of transparent object is important to consider for realistic rendering of transparent objects. However, light refraction is influenced by the index of refraction (IOR) and geometry of transparent objects. Furthermore, the refraction varies along the viewpoint, leading to the changes in the object's appearance. For the reasons, novel view synthesis of transparent objects based on color images is a challenging task.

Existing works on novel view synthesis include transparent object reconstruction and neural rendering.
To reconstruct transparent objects, special hardware setups like light field probes \cite{6126367}, turntables \cite{10.1145/3414685.3417815} have been required. Recently, Li \textit{et al.} \cite{li2020through} propose neural network that obtains geometry of transparent objects using a few images. However this method has the constraint of requiring an environment map. Also, when presented with an image that significantly deviates from the training data, it often produces inaccurate estimates. 
The latter, neural rendering, has seen significant advancements. In particular, Neural Radiance Fields (NeRF) \cite{mildenhall2021nerf} has gained attention as a new approach in computer graphics and computer vision, offering high-quality novel view synthesis. NeRF generates 3D representations of scene from 2D images by using machine learning and volume rendering. 
NeRF uses multi-layer perceptrons (MLPs) which take 3D coordinates (x, y, z) of sample points on light ray and viewing direction ($\theta, \phi$) indicating the direction of the ray as inputs and estimating the density and color values at those coordinates. 
The density and color values are used to predict pixel value by applying volume rendering. 
NeRF does not consider the refraction of rays and uses only straight rays. 
NeRF samples points along straight rays, leading to a challenge in accurately representing scenes containing transparent objects where ray refraction occurs.
Recent studies have proposed various approaches to address the problem. Bemana \textit{et al.} \cite{bemana2022eikonal} mark the area corresponding to transparent objects with a bounding box and optimize the 3D spatially-varying IoR inside the box. The curved way of light paths is determined by an eikonal equation that calculates based on the spatial gradients of the IoR field. However, bounding box  approximately encompasses the area corresponding to transparent object, making it challenging to synthesize scene accurately for complex-shaped transparent objects. Fujitomi \textit{et al.} \cite{9897642} propose the offset field network which estimates an offset from the straight line starting at the camera center. 
This allows the network to train without prior knowledge such as IOR or the shape of the object. However, due to the implicit representation of light refraction, it can not handle complex light paths.
\begin{figure*}[!t]
  \centering
  \includegraphics[width=\textwidth]{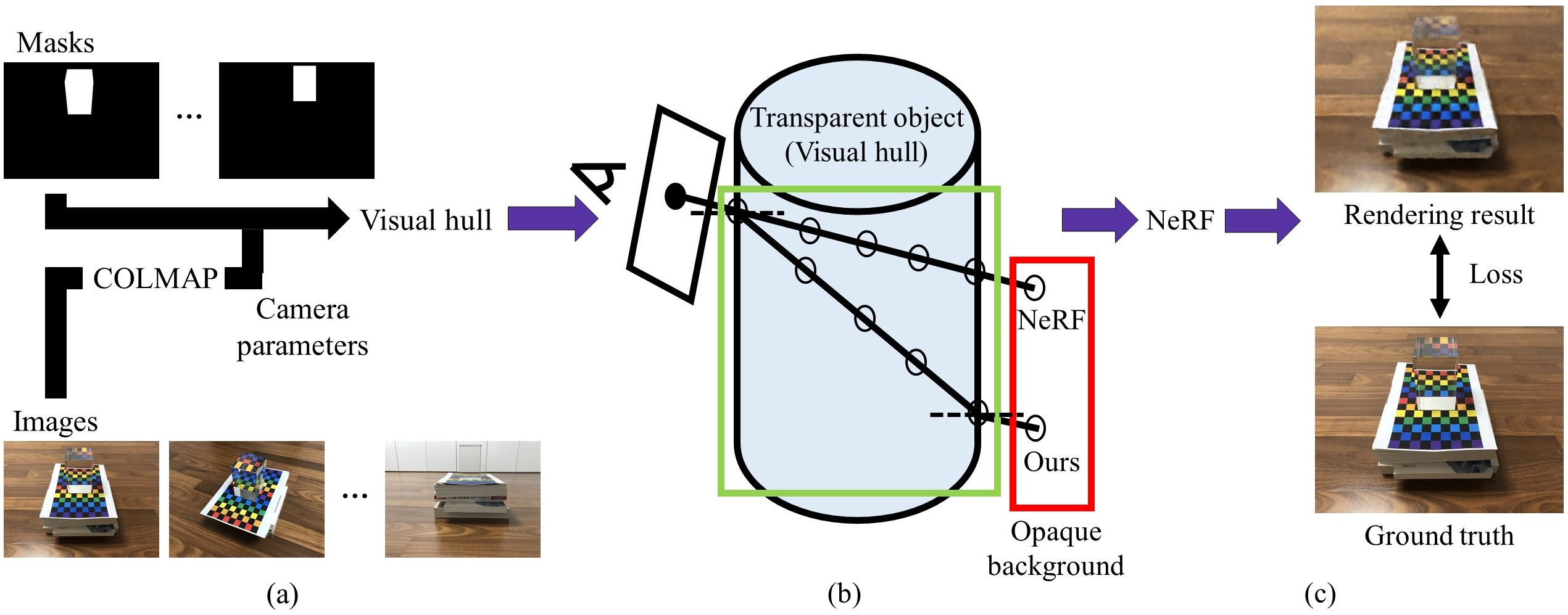}
  \caption{
  Proposed method (a) Given multi-view images, we use COLMAP \cite{7780814} to estimate camera parameters. We then create a 3D transparent object using visual hull which takes masks and camera parameters as input. (b) Comparison between NeRF (only straight rays), and our approach (allow refracted rays). Light refraction is calculated based on surface normal of reconstructed object, applying Snell's Law. We sample points along the refracted path of light. (c) We let NeRF estimate color and density of each point, and render pixel color. Finally, we get loss between the rendered and ground truth colors.
  }
  \label{fig:1}
\end{figure*} 

In this paper, we tackle the problem of novel view synthesis for transparent objects, which involves requiring complex hardware setup or absense of geometry. To end this, we propose a method based on NeRF, mathematical equations for the description of light refraction, and visual hull \cite{273735}, which reconstructs 3D volume from intersecting cones created by back projection of object silhouettes observed from multiple viewpoints. 
The reason why three concepts came out is as follows. NeRF shows outperforming performance in terms of view synthesis compared to existing models \cite{mildenhall2019local, sitzmann2019scene}. However, because NeRF uses a straight ray, it is not suitable for synthesis a scene with transparent objects where light is refracted. To calculate light refraction, we use mathematical equations which require object's geometry. So we get 3D object through visual hull reconstruction. One of the advantages of visual hull is its ease of transparent object reconstruction with minimal requirements, as it can be done with camera parameters and masks, without the need for complex setup. Camera parameters can be obtained easily through COLMAP \cite{7780814}, and masks also can be acquired conveniently using semantic segmentation method \cite{he2021enhanced}. 
To summarize, We use multi-view images of a transparent object with known IOR. 
We start with visual hull to obtain 3D volume of target object.
And then we correct light ray that refracts as it passes through a transparent object using Snell’s law. With the corrected new light path, we sample points along that path and feed them to NeRF.

\section{Method}
\subsection{Transparent Object Reconstruction}
We use visual hull for reconstructing transparent object. The conventional 3D reconstruction method, Structure-from-Motion, is vulnerable to color discordance on the surface of transparent objects,
making the 3D reconstruction of transparent objects unreliable. On the other hand, visual hull reconstruct an object regardless of color variation as long as we obtain reliable silhouettes(masks) of the object observed from multiple view. Therefore, we employ visual hull to obtain 3D shape of transparent object. 
\\
We take N multi-view images with known IOR and run COLMAP to obtain camera parameters. We choose M images, using them as input into the semantic segmentation model \cite{he2021enhanced} to obtain masks. \cite{he2021enhanced} propose two modules for solving glass-like object segmentation. one module for predicting object boundary and the other module for overall shape of glass objects representing along the boundary. Next, we use selected M images to compute image projection matrices. After that, we project 3D volume with a size of $K^3$ voxels onto the image planes. To reconstruct the 3D shape, we preserve voxels that consistent with the masked regions within the masks and subtract voxels that inconsistent. Then we use marching cubes to get visual hull and apply smoothing.

\subsection{Point Sampling on Refracted Ray}
In this step, there are 2 assumptions to simplify the problem. First, when light encounters a transparent object, it splits into two paths through reflection and transmission. We consider only the transmitted light. Second, we include maximum two light bounces occurring (front and rear surface of a transparent object) as shown in Fig. 1. (b).
\\
We conduct ray casting to investigate which ray intersects the front surface of the transparent object reconstructed by visual hull at point x. In order to calculate the direction of the refracted ray at point x, we use the following equations. And then we once again apply ray casting and equations for the rear surface. After obtaining the refraction of light in this manner, we uniformly sample points along the path of this refracted ray.

\begin{equation}
\label{eqn:01}
\cos\theta_{1} = -\vec{n}\cdot\vec{l},
\end{equation}
\begin{equation}
\label{eqn:02}
\cos\theta_{2} = \sqrt{1-\left(\frac{n_{1}}{n_{2}}\right)^2(1-(\cos\theta_{1})^2)},
\end{equation}

\begin{subnumcases}{\vec{v} = }
    \vec{l} + 2\vec{n}(-\vec{l}\cdot\vec{n})    & \text{if $cos\theta_{2}$ $<$ 0}. \\[10pt]
    \frac{n_{1}}{n_{2}}\vec{l} + (\frac{n_{1}}{n_{2}}\cos\theta_{1} - \cos\theta_{2})& \text{otherwise}
\end{subnumcases}
where $\vec{n}$ is normalized surface normal vector, $\vec{l}$ is normalized incoming vector, $n_{1}$ is the refractive index of the medium through which light enters, $n_{2}$ is the refractive index of a medium that light leaves, $\theta_{1}$ is the angle of incidence, $\theta_{2}$ is the angle of refraction, and $\vec{v}$ is refracted ray.
Equation (3a) considers total internal reflection which happen only when a ray penetrates less-dense medium ($n_{2}$ $<$ $n_{1}$). In other words, When ray exits a transparent object and the angle of incidence exceeds the critical angle, reflection occurs.

\subsection{Network Optimization using Volume Rendering}
\subsubsection{Volume Rendering}
To represent scene, we need to consider two types of ray: ray that is refracted when they hit transparent object and ray that continues straight to the background without interacting with the transparent object.
At each point sampled uniformly from these rays, network estimates color and density values and the pixel color is determined using volume rendering which is same equation as NeRF.

\begin{equation}
\label{eqn:04}
    \begin{split}
    \hat{C}(r) &= \sum\limits_{i=1}^N T_{i}(1-exp(-\sigma_{i}\delta_{i}))c_{i}, 
    \\
    \text{where } T_{i} &= exp(-\sum\limits_{j=1}^{i-1} \sigma_{i}\delta_{i})
    \end{split}
\end{equation}
where $\hat{C}(r)$ is rendered pixel color of the ray r, $c_{i}$ and $\sigma_{i}$ are the color and density values estimated by the network at the $i$th sampling point, and $\delta_{i}$ is the distance between the $i$th and ($i+1$)th. 

\subsubsection{Loss Function}
We optimize MLPs using the following equation which calculates the loss between rendered and true pixel colors.
\begin{equation}
\label{eqn:04}
    \mathcal{L} = \frac{1}{V}\sum\limits_{i=1}^V ||\hat{C}(r_{i})-C(r_{i})||_{2}^2
\end{equation}
where $C(r)$ is true pixel color and $V$ is the number of rays in each batch.

\section{Results and Discussion}
\subsection{Datasets}
We evaluate our method on three datasets: Cubic, WaterBottle, and PlasticCup. Cubic is made of glass. WaterBottle and PlasticCup are filled with water. Both datasets are real scenes containing transparent object with opaque background. we set the IOR for each object to 1.5, 1.4, and 1.35. In each dataset, we capture 84, 102, and 81 images with an iPhone 7 Plus and used 10\% of them as test set. Total number of selected images needed for obtaining masks which are input of visual hull is 8. The resolution of all the images is 672 x 504. 

\subsection{Implementation Details}
We set the size K of the 3D volume required to obtain the visual hull to 256. 
We use batch size of 1024 rays and set the number of sampling points along the ray to 128. All models are run for 80k iterations.

\subsection{Qualitative Comparison}
We compare our method with NeRF and results are shown in Fig. 2. Our method reasonably renders distorted opaque background (color checkerboard pattern) caused by transparent objects. Rendered background pattern is more clear and sharp compared to the blurry result of NeRF. 
\\
However, in both WatterBottle and PlasticCup, curved surfaces are rendered with some blurriness. Visual hull we used is not capable of accurately modeling curved surfaces since it carves out the 3D volume to reconstruct the object's shape. As a result, the normal vector on the surface is not precise, which in turn affects the light's refractive path using Snell's law. In addition, top surface of Cubic and water surface on PlasticCup have unclarity. These refracted patterns appear with more than two light bounce occurrences, and we lack the ability to express the details of those surfaces because we consider the light bounce up to two times.

\begin{figure}[!t]
  \centering
  \includegraphics[width=\linewidth]{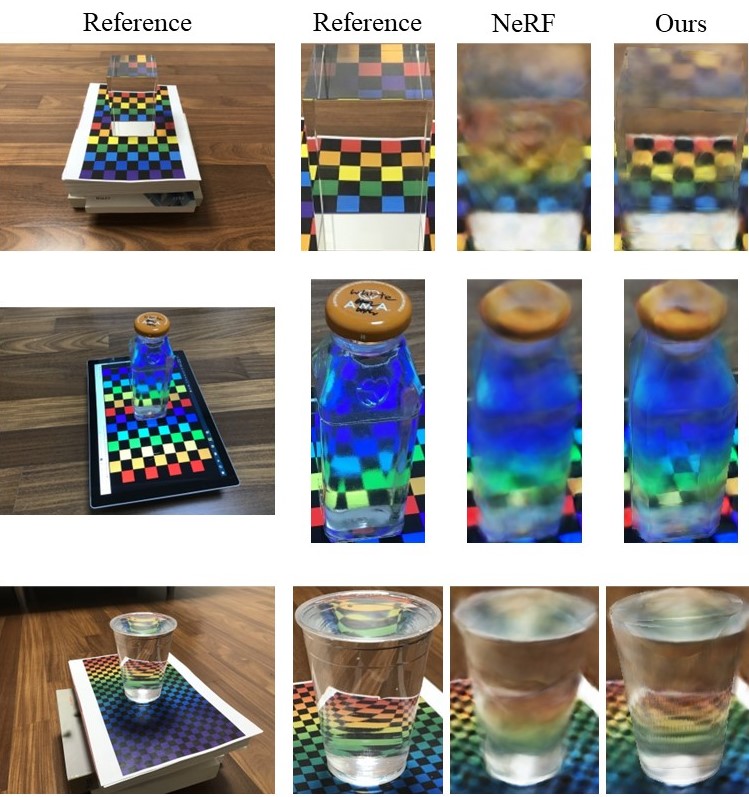}
  \caption{
  Qualitative comparison of novel view synthesis for real scenes
  }
  \label{fig:2}
\end{figure} 

\subsection{Straight and Refracted rays}
We select one pixel, (310, 125), from the pixels within the transparent object region. Fig. 3 shows graphs depicting the sampling points of ray corresponding to the selected pixel from both NeRF which uses straight ray and our method which uses refracted ray. Green and red boxes indicate each area shown in Fig. 1. (b). Green box indicates inside the transparent object, and red box indicates checkerboard of background.
\\
NeRF estimates that the density of points in the green box area is high. The model learn that there is an opaque object in that area, not a transparent object. As shown in Fig. 3. (a), estimated density values of 15th to 20th points are high, incorrect color values is entered in the volume rendering process, and as a result, the rendered image looks noisy and blurry.
In addition, because a straight ray is used, the ray of the pixel touches the black area on the checkerboard, as a result, the network estimates the color value in black, like the color of the point in Fig. 3. (a) red box area. 
However, Fig. 3. (b) shows that our method estimates lower density in the green area than NeRF indicating that it is a transparent object region. Therefore, we have a better ability to generate clear rendered images.
Moreover, by using refracted light, the color value of the points in the red box are estimated to be red consistent with truth.
\begin{figure}[!t]
  \begin{minipage}[c]{\linewidth}
  \subcaptionbox{NeRF's sampling points on a straight ray}
  {\includegraphics[width=1\columnwidth]{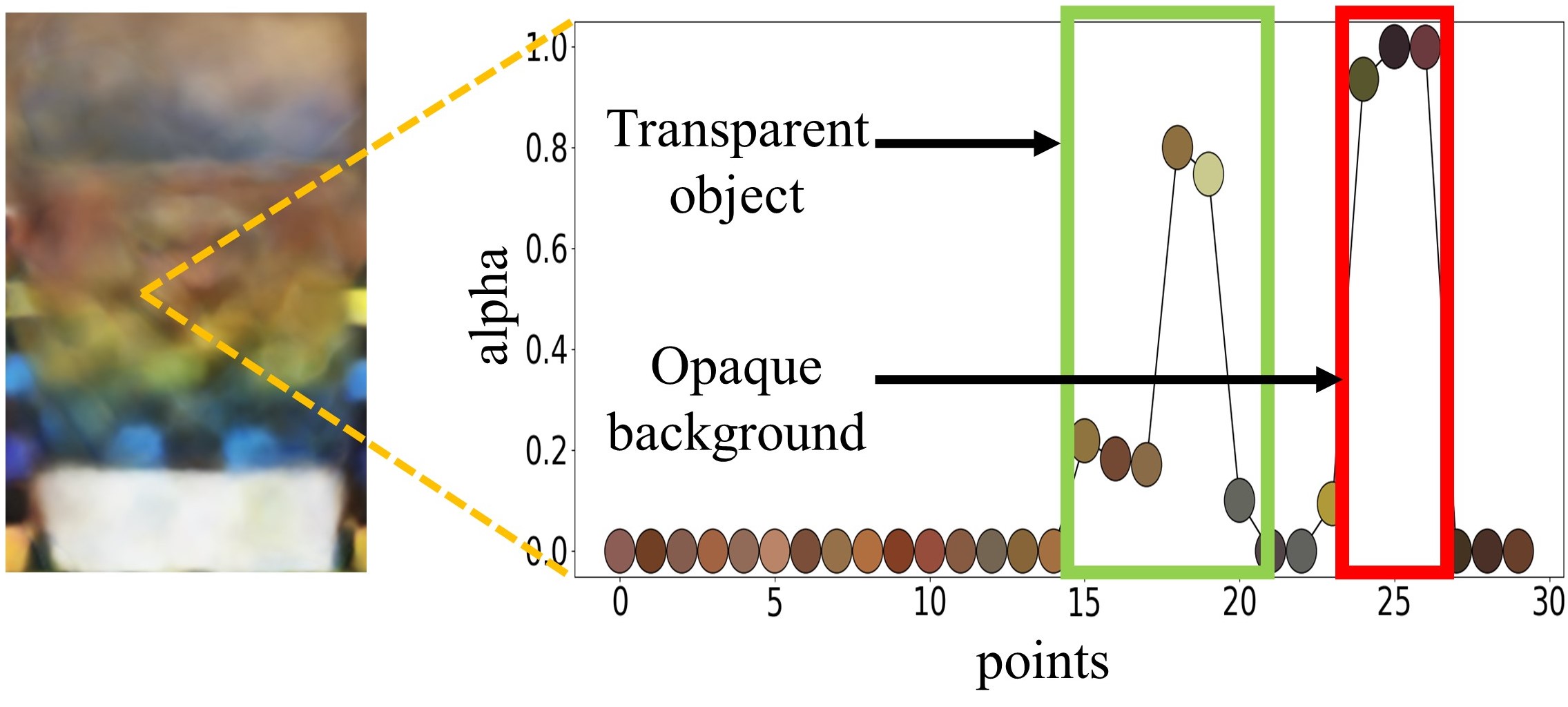}}\end{minipage}%
  \vspace{0.5cm}
  \begin{minipage}[c]{\linewidth}
  \subcaptionbox{Our method's sampling points on a refracted ray}
  {\includegraphics[width=1\columnwidth]{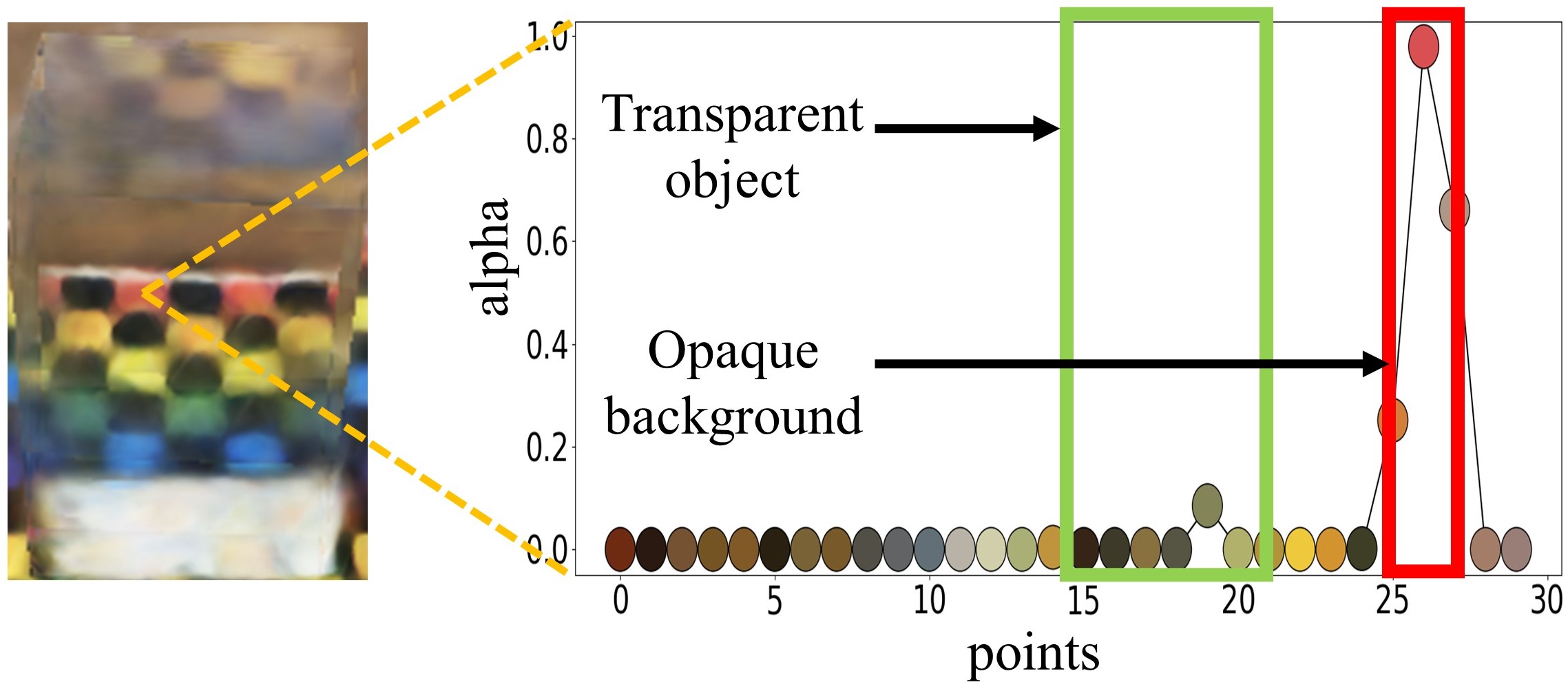}}\end{minipage}%
  \caption{
  Ray comparison of of NeRF and ours. Our method shows better estimation of background color values and transparent object's density values compared to NeRF.
  }
  \label{fig:3}
\end{figure} 

\section{Conclusion}
Prior approaches to novel view synthesis for transparent objects had limitations. They either require complex equipment or are suitable for only simple-shaped objects due to missing explicit estimation of light refraction using object geometry. To resolve these problems, We propose the following method using multi-view images. We obtain a 3D volume of the target object and use mathematical equations to describe light refraction. After that, we sample points along with refracted light and apply them to NeRF. As a result, we can increase clarity and enhance surface details.
\\
Our approach has limitations. First, reconstruction of transparent object using visual hull is vulnerable to curved surfaces. This lead to generating blurry results. With the applying of neural SDF(Signed Distance Function) \cite{yariv2020multiview}, we can improve quality of rendered images. Second, we consider only transmission occurs on a object's surface. We believe that more realistic image which contains reflection on a surface can be synthesized by using Fresnel equation. Lastly, we set the number of light bounce that occur in transparent object to 2. We can handle more complex light path by using  differentiable path tracing.

\bibliographystyle{IEEEtran}
\bibliography{IEEEabrv,ref}




\end{document}